\documentclass[iicol,sn-basic]{sn-jnl}

\usepackage{graphicx}%
\usepackage{multirow}%
\usepackage{amsmath,amssymb,amsfonts}%
\usepackage{amsthm}%
\usepackage{pifont}

\usepackage{mathrsfs}%
\usepackage[title]{appendix}%
\usepackage{xcolor}%
\usepackage{textcomp}%
\usepackage{manyfoot}%
\usepackage{booktabs}%
\usepackage{algorithm}%
\usepackage{algorithmicx}%
\usepackage{algpseudocode}%
\usepackage{listings}%

\DeclareRobustCommand{\method}{{PAT}}

\DeclareRobustCommand{\char}{Charades}
\DeclareRobustCommand{\thum}{MultiTHUMOS}
 
\newcommand{\cm}{\textcolor[rgb]{0.49,0.49,0.49}} 
\DeclareRobustCommand{\bcolor}{\textcolor[rgb]{0.0,0.1,0.95}}

\DeclareRobustCommand{\eg}{\textit{e.g.}}

\DeclareRobustCommand{\ie}{\textit{i.e.}}

\theoremstyle{thmstyleone}%
\theoremstyle{thmstyletwo}%

\theoremstyle{thmstylethree}%

\begin{document}

\title[Article Title]{An Effective-Efficient Approach for Dense Multi-Label Action Detection}



\author*[1]{\fnm{Faegheh} \sur{Sardari}}\email{f.sardari@surrey.ac.uk}

\author[1]{\fnm{Armin} \sur{Mustafa}}\email{armin.mustafa@surrey.ac.uk}

\author[1]{\fnm{Philip J. B.} \sur{Jackson}}\email{p.jackson@surrey.ac.uk}

\author[1]{\fnm{Adrian} \sur{Hilton}}\email{a.hilton@surrey.ac.uk}

\affil[1]{\orgdiv{Centre for Vision, Speech and Signal Processing (CVSSP)}, \orgname{University of Surrey}, \country{UK}}



\abstract{Unlike the sparse label action detection task, where a single action occurs in each timestamp of a video, in a dense multi-label scenario, actions can overlap. To address this challenging task, it is necessary to simultaneously learn (i) temporal dependencies and (ii) co-occurrence action relationships. Recent approaches model temporal information by extracting multi-scale features through hierarchical transformer-based networks. However, the self-attention mechanism in transformers inherently loses temporal positional information. We argue that combining this with multiple sub-sampling processes in hierarchical designs can lead to further loss of positional information. Preserving this information is essential for accurate action detection. In this paper, we address this issue by proposing a novel transformer-based network that (a) employs a non-hierarchical structure when modelling different ranges of temporal dependencies and (b) embeds relative positional encoding in its transformer layers. Furthermore, to model co-occurrence action relationships, current methods explicitly embed class relations into the transformer network. However, these approaches are not computationally efficient, as the network needs to compute all possible pair action class relations. We also overcome this challenge by introducing a novel learning paradigm that allows the network to benefit from explicitly modelling temporal co-occurrence action dependencies without imposing their additional computational costs during inference. We evaluate the performance of our proposed approach on two challenging dense multi-label benchmark datasets and show that our method improves the current state-of-the-art results by 1.1\% and 0.6\% per-frame mAP on the {\char} and {\thum} datasets, respectively, achieving new state-of-the-art per-frame mAP results at 26.5\% and 44.6\%, respectively. We also performed extensive ablation studies to examine the impact of the different components of our proposed network and learning paradigm.}

\keywords{Dense multi-label action detection, position-aware transformer, temporal multi-scale features, co-occurrence class action relations}



\maketitle

\section{Introduction}
\label{sec:intro}
\begin{figure}[t]
  \centering
  \includegraphics[width=1.0\linewidth]{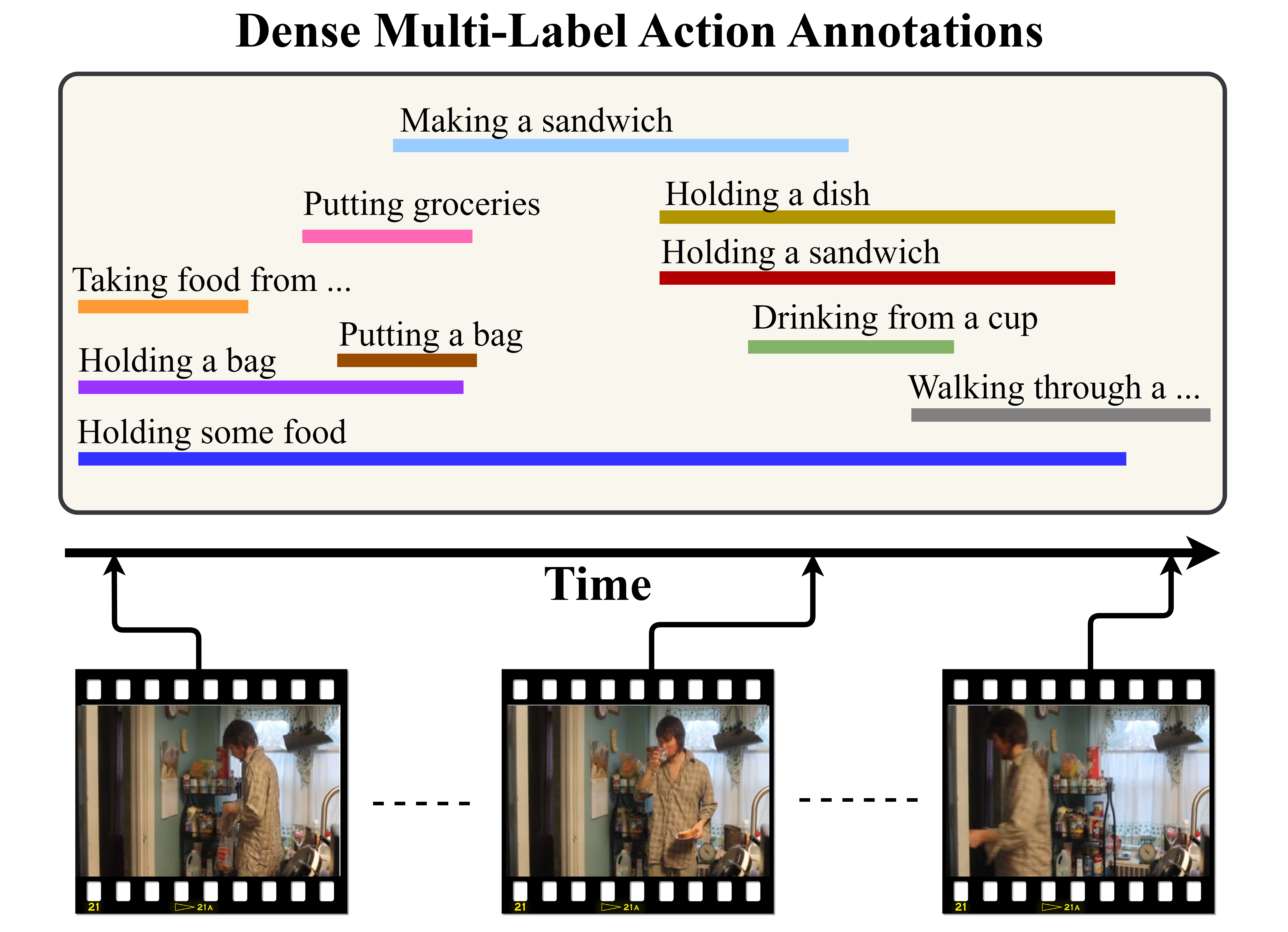}
  \caption{A sample video and its corresponding action annotations from the {\char} dataset (\cite{charades}) where the video includes several action types with different time spans, from short to long, and in each time step, multiple actions can occur at the same time.\vspace{1.5mm}}
  \label{fig:dense detection}
\end{figure}

Action or event detection aims to determine the boundaries of different actions/events occurring in an untrimmed video and plays a crucial role in various important computer vision applications, such as video summarization (\cite{jiang2022joint, wu2022intentvizor,he2023align}), video highlighting (\cite{badamdorj2021joint,wei2022learning,moon2023query}), and video captioning (\cite{Wang_2021_ICCV,yang2023vid2seq,Nadeem_2023_CVPR,nadeem2024cad}). Despite the recent advances in different areas of video understanding, dense multi-label action detection is still an unsolved problem and is considered as one the most challenging video analysis tasks since the videos are untrimmed, {\ie} a video's length can range from seconds to several minutes, {and include several actions/events that have different ranges of time duration}. Furthermore, multiple actions can overlap in the video, {\eg}, in the {\char} dataset (\cite{charades}), you can find sample videos that 12 actions occur at the same time. {Fig. \ref{fig:dense detection} shows a video sample of {\char} benchmark dataset and its corresponding action annotations.} To carry out this challenging task, we require to simultaneously learn (i) {\bf temporal relationships} and (ii) {\bf co-occurrence action dependencies}.  
Most previous dense multi-label action detection approaches capture these relationships through temporal convolutional networks (\cite{piergiovanni2019temporal, feichtenhofer2020x3d, kahatapitiya2021coarse}). However, with the success of transformer networks over the convolutional networks in modeling complex and sequential dependencies (\cite{vaswani2017attention}), recently, a few methods, such as (\cite{tirupattur2021modeling, dai2021ctrn, dai2022ms}), leverage the self-attention mechanism and propose transformer-based approaches.

Although the transformer-based action detection approaches achieve state-of-the-art performance, they suffer from two main challenges. 
First, to model {\bf temporal relationships}, they commonly extract and combine multi-scale temporal features through hierarchical structures, such as (\cite{ zhang2022actionformer, dai2022ms}). In these approaches, their network contains several transformer layers such that the output of each layer is down-sampled and given as input into its subsequent layer. However, as stated in (\cite{shen2018disan, li2021learnable, dufter2022position}), the self-attention mechanism in the transformer is order-invariant and loses positional information. We argue that when the self-attention is embedded in a hierarchical structure, the issue becomes worse as using multiple down-sampling processes results in {increased loss of} positional information, especially in top layers, while preserving this information is crucial for accurate action detection. Second, to model {\bf co-occurrence action relations}, they embed cross-action dependencies into the transformer layers (\cite{tirupattur2021modeling, lin2021exploring}). However, these approaches are not computationally efficient, particularly when the number of action classes are large as the network needs to compute all possible pair action class relations. For example, in (\cite{tirupattur2021modeling}), to investigate action relationships, they map each timestamp of the video into the maximum number of action classes, and apply a transformer layer on the actions classes of each timestamp individually. Therefore, for a video of length $T$ and $C$ action classes, they require to compute $T\times C\times C$ connections.


In this paper, we overcome the above challenges by proposing a novel transformer-based network, trained through an innovative learning paradigm. Our proposed network relies on two transformer-based branches: Assistant and Core. During training, both the video data and its corresponding ground-truth labels serve as inputs to our network. The Assistant branch receives the ground-truth labels to explicitly learn co-occurrence action class relationships through a self-supervised autoencoder technique. In contrast, the Core branch takes the video to perform dense multi-label action detection by exploiting multi-scale temporal video features while designed to preserve positional information through (i) a novel non-hierarchical structure and (ii) embedding relative positional encoding (\cite{shaw2018self}) in its transformer layers. In the Core branch, action class relations are not explicitly modelled; however, they are learned implicitly through interaction with the Assistant branch and our proposed learning paradigm. During inference, only the Core branch is deployed for the dense action detection task. This approach allows the network to benefit from explicitly modelling co-occurrence action relationships during training without imposing their computational costs at inference. {Our key contributions are summarized as follows: 
\begin{itemize}
     \item We pioneer the idea of leveraging positional information in transformers for action detection.
     \item We introduce a novel learning paradigm that allows the network to explicitly leverage co-occurrence action class relationships during the training without adding their computational costs at the inference - this can benefit the training process in other transformer-based networks. 
    \item  We propose a novel {non-hierarchical} transformer-based network that preserves positional information while exploiting multi-scale temporal dependencies.
    \item We perform extensive ablation studies to evaluate {our network design} and learning paradigm.
    \item We evaluate the performance of our proposed approach on two challenging benchmark dense action detection datasets, {\char} and {\thum}, where we outperform the current state-of-the-art by $1.1\%$ and $0.6\%$ per-frame mean Average Precision (mAP) on {\char} and {\thum} respectively, thereby achieving the new state-of-the-art per-frame mAP at $26.5\%$ and $44.6\%$, respectively.
    
\end{itemize}


\section{Related Works}
\label{sec:related works}
Although action detection (\cite{chao2018rethinking, lin2019bmn,xu2020g,ma2020sf, lin2021learning, chang2021augmented, zhang2022actionformer, vahdani2022deep, shi2023tridet}) has been studied significantly in computer vision, few works (\cite{piergiovanni2018learning,dai2019tan, piergiovanni2019temporal, dai2021pdan, kahatapitiya2021coarse}) have explored it in a dense multi-labelled setup where instances of different actions or events can overlap in different parts of a video. In this section, we review the action detection approaches by focusing on a dense-labelled setting. 

To detect the boundaries of different actions, the authors in (\cite{chao2018rethinking,lin2018bsn,liu2019multi,li2021three}) propose anchor-based methods where they first generate several proposals for each frame of video by using multi-scale anchor boxes, and then refine them to achieve the final action boundaries. However, these approaches are not usually applied for a dense multi-label scenario, as to model effectively the dense action distributions, they need a large amount of anchors. To overcome this, some works, such as (\cite{piergiovanni2018learning,dai2019tan, piergiovanni2019temporal, dai2021pdan, kahatapitiya2021coarse}), design anchor-free approaches for dense action detection. \citet{piergiovanni2018learning} propose a network that represents an untrimmed video into multi-activity events. They design multiple temporal Gaussian filters which are applied separately on the video frame features while a soft-attention mechanism is employed to combine the output of the filters to generate a global representation. Later in (\cite{piergiovanni2019temporal}), they improve their work by proposing a temporal convolutional network using Gaussian filters as kernels to perform the temporal representation in a more efficient and effective way. Although they design networks to address complex multi-label action detection, the proposed models are not able to encode long-term dependencies and mostly focus on local relationships. In contrast, the Core branch of our proposed network has been designed to capture different ranges of temporal dependencies from short to long.  \citet{kahatapitiya2021coarse} propose a two-stream network to capture long term information such. In their network, one of the streams learns the most informative frame of a long video through a dynamic sub-sampling with a ratio of 4, and the other one exploits the fine-grained contexts of the video from the full resolution. Although their results are promising, their method cannot be adapted easily to use more temporal resolutions as it requires a dedicated Convolutional Neural Network (CNN), \ie, X3D (\cite{feichtenhofer2020x3d}), for each resolution, {whereas in our proposed method, a different resolution can be processed easily by adding an extra branch containing a few transformer blocks in the Core branch.}

\vspace{2mm}
\noindent{{\bf Transformer-based Approaches -- }}
With the success of transformer networks in modeling complex relationships and capturing short and long term dependencies, some approaches, such as (\cite{dai2021pdan, tirupattur2021modeling, dai2022ms}), develop transformer-based networks for dense action detection task. \citet{tirupattur2021modeling} design a model with two transformer components, one component applies self-attention across all action classes for each time step to learn the relationships amongst actions, and another component uses self-attention across {time frames} to model the temporal dependencies, and the output of two components are combined for action classification.  Although this method outperforms state-of-the-art results, it is not computationally efficient as its computational complexity relies on the number of action classes.  In contrast, our proposed network and  learning paradigm allows us to benefit from explicitly learning class-action relations during the training without incurring their computational costs during inference.  Similar to (\cite{kahatapitiya2021coarse}) that benefits different temporal resolutions, \citet{dai2022ms} extract multi-scale features. They design a transformer-based hierarchical structure and provide multi-resolution temporal features through several sub-sampling processes. However, as the self-attention mechanism does not preserve the temporal position information (\cite{shen2018disan, li2021learnable, dufter2022position}), joining it with multiple sub-sampling processes makes the network lose more positional information. However, preserving this information is essential for action detection.  In contrast, in our proposed approach, the Core branch has been designed to retain positional information while learning multi-scale temporal features.

\section{Method}
\label{sec:proposed method}
{In this section, we first define dense multi-label action detection task. Then, we elaborate on our proposed network and learning paradigm.} 

\subsection{Problem Definition} 
In dense multi-label action detection task, we aim to detect all actions/events happening in each timestamp of an untrimmed video as defined in (\cite{kahatapitiya2021coarse, tirupattur2021modeling, dai2022ms}}). For an untrimmed video sequence with a length of $T$, each timestamp $t$ has a ground truth action label ${G}_t =\{{{g}_{t,c}}\in\{0, 1\}\}^{C}_{c=1}$, where $C$ is the maximum number of action classes in the dataset, and the network requires to estimate action class probabilities $Y_t =\{y_{t,c}\in[0, 1]\}^{C}_{c=1}$ for each timestamp.

\subsection{Proposed Network \& Learning Paradigm} 
In this section, we introduce a novel transformer-based network that is trained through an innovative learning paradigm. Our approach is designed to effectively and efficiently learn complex temporal co-occurrence action relationships for dense multi-label action detection. 

\subsubsection{Proposed Network}  
\begin{figure}[t]
  \centering
  \includegraphics[width=1.0\linewidth]{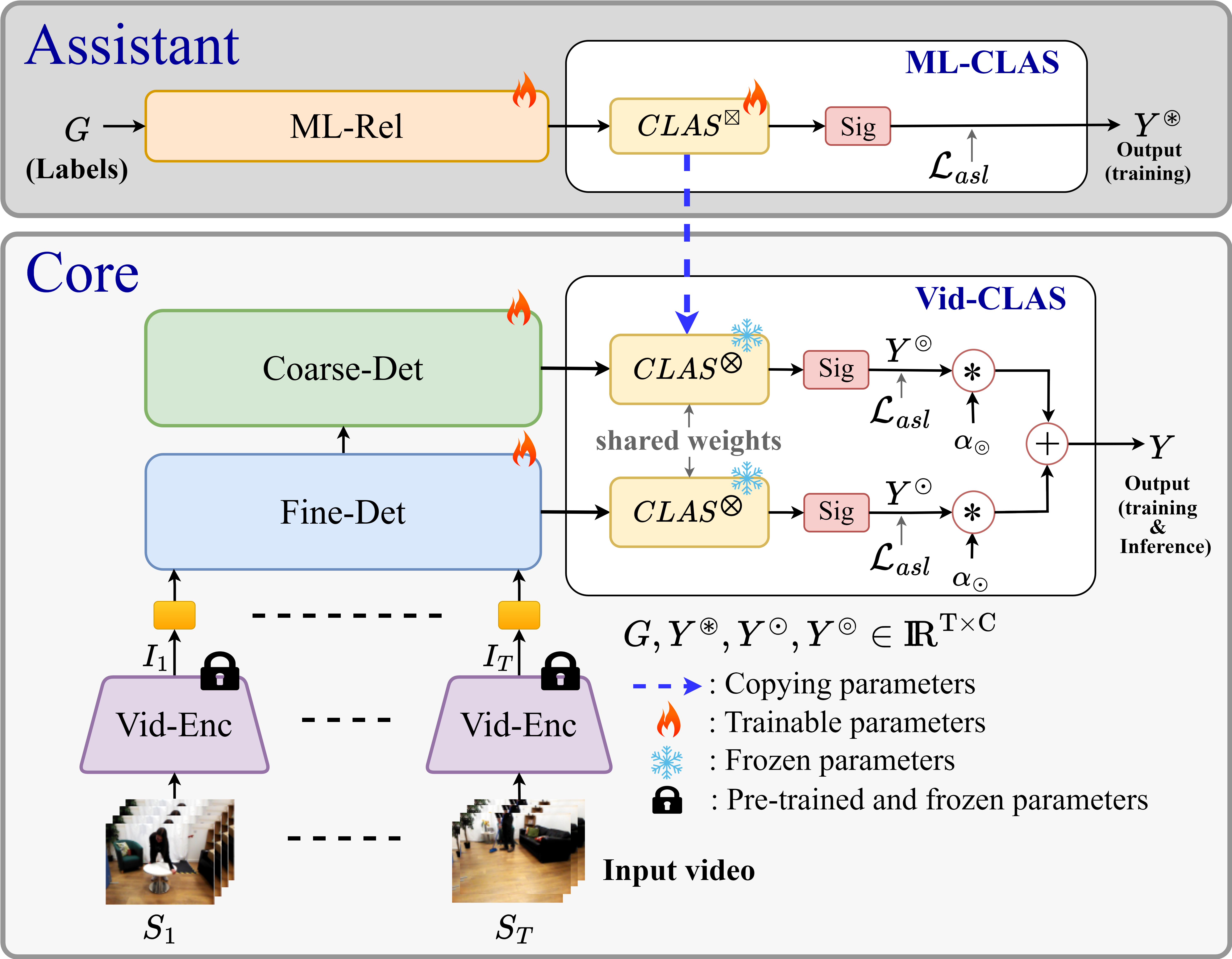}
  \caption{The overall schema of our proposed network that includes two branches: Assistant and Core. The Assistant branch comprises the multi-label relationship (ML-Rel) and multi-label classification (ML-CLAS) modules. The Core branch consists of a video encoder (Vid-Enc) and three main modules: fine detection (Fine-Det), coarse detection (Coarse-Det), and video classification (Vid-CLAS). During training, both the Assistant and Core branches are employed, while at the inference, only the Core branch is deployed.} 
  \label{fig:MF-ViT}
\end{figure}

Our network comprises two transformer-based branches: Assistant and Core, as illustrated in Fig. \ref{fig:MF-ViT}. During training, both the video data and its corresponding ground-truth labels serve as inputs to our network. The Assistant branch receives the ground-truth labels to explicitly model co-occurring action class relationships. In contrast, the Core branch receives the video data to perform the dense multi-label action detection task by exploiting multi-scale temporal video features. The Core branch does not explicitly model action relationships; however, these relationships are learned implicitly through interaction with the Assistant branch and our proposed learning paradigm. During inference, only the Core branch is deployed for action detection. This design enables the network to benefit from explicitly modelling cross-action relationships during training without incurring their computational cost at inference.

\vspace{2mm}
\noindent{\bf{Assistant Branch -}} The Assistant branch learns co-occurrence relationships among action classes by employing a self-supervised autoencoder technique on the video's ground-truth labels. To achieve this, it consists of a multi-label relationship ML-Rel module followed by a multi-label classification ML-CLAS module (see Fig. \ref{fig:MF-ViT}). The ML-Rel module utilizes the self-attention mechanism to encode action class relationships from the video's ground-truth labels, while the ML-CLAS module designed based on the 1D convolution filter decodes the learned action class relationships to the original input labels.


\vspace{2mm}
\noindent{\bf{Core Branch -}} The  Core branch detects dense actions from the input video data by relying on a video encoder Vid-Enc and three key components: fine detection Fine-Det, coarse detection Coarse-Det, and video classification Vid-CLAS modules (see Fig. \ref{fig:MF-ViT}). In the Core branch, first, Vid-Enc, a pre-trained model, encodes the input video into a sequence of input tokens. Then, the Fine-Det module, a transformer-based block, receives the input video tokens and processes them at their original temporal resolution to obtain a fine-grained action representation for both the Coarse-Det and Vid-CLAS modules. Subsequently, the Coarse-Det module, also a transformer-based block, exploits different ranges of temporal action dependencies among the fine-grained features by extracting and combining multi-scale temporal features. Finally, the Vid-CLAS module, designed based on the 1D convolution filter, estimates action class probabilities from the output of both Fine-Det and Coarse-Det modules.


\vspace{2mm}
\noindent{In contrast to previous transformer-based action detcetion approaches, such as (\cite{zhang2022actionformer, dai2022ms, tirupattur2021modeling}), which lose temporal positional information when modeling temporal dependencies, our network is designed to leverage this essential information in its transformer layers. To achieve this, it applies a non-hierarchical structure to extract multi-scale temporal features in the Core branch. Additionally, to further preserve positional information, it embeds relative positional encoding (\cite{shaw2018self}) in the transformer blocks of both the Core and Assistant branches.}

\subsubsection{Proposed Learning Paradigm} Our proposed learning paradigm aims to encourage the Core branch to learn the same co-occurrence action class relationships as those learned from the video's ground-truth labels in the Assistant branch. The workflow of our learning paradigm is presented in the Algorithm \ref{alg: our algorithm}.

\begin{algorithm}
\caption{Our proposed learning paradigm.}\label{alg:cap}
\label{alg: our algorithm}
{{\bf{Training set:}} a set of video sequences $V$ and its ground-truth labels $A$, where $V=\{V_i\}^{N}_{i=1}$ and $A=\{A_i\}^{N}_{i=1}$}

K: total epoch numbers
\begin{algorithmic}
\For{$k=1$ to $K$}
\For{each $\{V_i, A_i\}$ in the training set}
\State Randomly select a T-frame video clip of 
\State $V_i$, $S = \{S_t\}^{T}_{t=1}$, and its  corresponding 
\State annotations in $A_i$, $G = \{G_t\}^{T}_{t=1}$ .
\State
\State \cm{\% Assistant Branch}
\State $\hat{G} = \text{ML-Rel}(G)$
\State $Y^{\circledast} = \text{ML-CLAS}(\hat{G})$
\State Compute loss $\mathcal{L}_{Assis}(Y^{\circledast},G)$
\State \bcolor{Backward pass}
\State 
\State \cm{\% Copying parameters}
\State {Vid-CLAS} $\gets$ {ML-CLAS}
\State
\State \cm{\% Core Branch}
\State ${I}= \text{Vid-Enc}(S)$
\State $I^{\odot}=\text{Fine-Det}(I)$
\State $I^{\circledcirc}= \text{Coarse-Det}(I^{\odot})$
\State \cm{\% Video Classification's parameters are}
\State \cm{frozen}
\State $Y= \text{Vid-CLAS}(I^{\odot}, I^{\circledcirc})$
\State Compute loss $\mathcal{L}_{Core}(Y,G)$ 
\State \bcolor{Backward pass}
\EndFor

\EndFor
\end{algorithmic}
\end{algorithm}

Our learning paradigm begins by forwarding a video's multi-label action annotations to the Assistant branch. Following a backward pass, the ML-CLAS module's parameters are transferred to Vid-CLAS module of the Core branch. Next, the Core branch receives the video corresponding to the labels to estimate the probabilities of different action classes for each input token. During this training stage, the Vid-CLAS module's parameters are frozen, having been copied from the ML-CLAS module. This setup encourages the Fine-Det and Coarse-Det modules in the Core branch to learn the same temporal co-occurrence class action dependencies as those learned from the ground-truth labels in the ML-Rel module of the Assistant branch.

\subsubsection{Details of Proposed Network}
\noindent {\bf Relative Positional Transformer (RPT) Block -} As stated in (\cite{shen2018disan, li2021learnable, dufter2022position}), the self-attention mechanism in transformers loses the order of temporal information. However, preserving this information is essential for action detection since we need to precisely localize events in a video sequence. To overcome this issue, \citet{vaswani2017attention} propose to add the absolute positional embedding to the input tokens. However, in our experiments, we observed that using the absolute positional embedding decreases the method’s performance significantly (see Section \ref{sec:ablation}). This has also been observed in (\cite{dai2022ms, zhang2022actionformer}). The decrease in performance may be attributed to breaking the translation-invariant property of the method. In action detection, we expect the proposed method to be translation-invariant, {\ie}, the network learns the same representation for the same video frames in two temporally shifted videos, regardless of how much they are shifted, while the absolute encoding can break this property as it adds different positional encodings to the same frames in the shifted video inputs. To prevent the performance degradation, current approaches, such as (\cite{dai2022ms, zhang2022actionformer}), avoid using absolute positional encoding. In our network, we instead address this challenge by leveraging relative positional encoding (\cite{shaw2018self}) in the transformer layers. The relative positional encoding employs a relative pairwise distance between every two tokens and is translation-invariant. In addition, in contrast to the absolute positional encoding that is applied to only the input tokens, the relative positional encoding is embeded in each transformer layer, so the positional information can pass into the subsequent layer and flow to the classification module where the final localisations are provided.

\begin{figure}[t]
  \centering
  \includegraphics[width=1.0\linewidth]{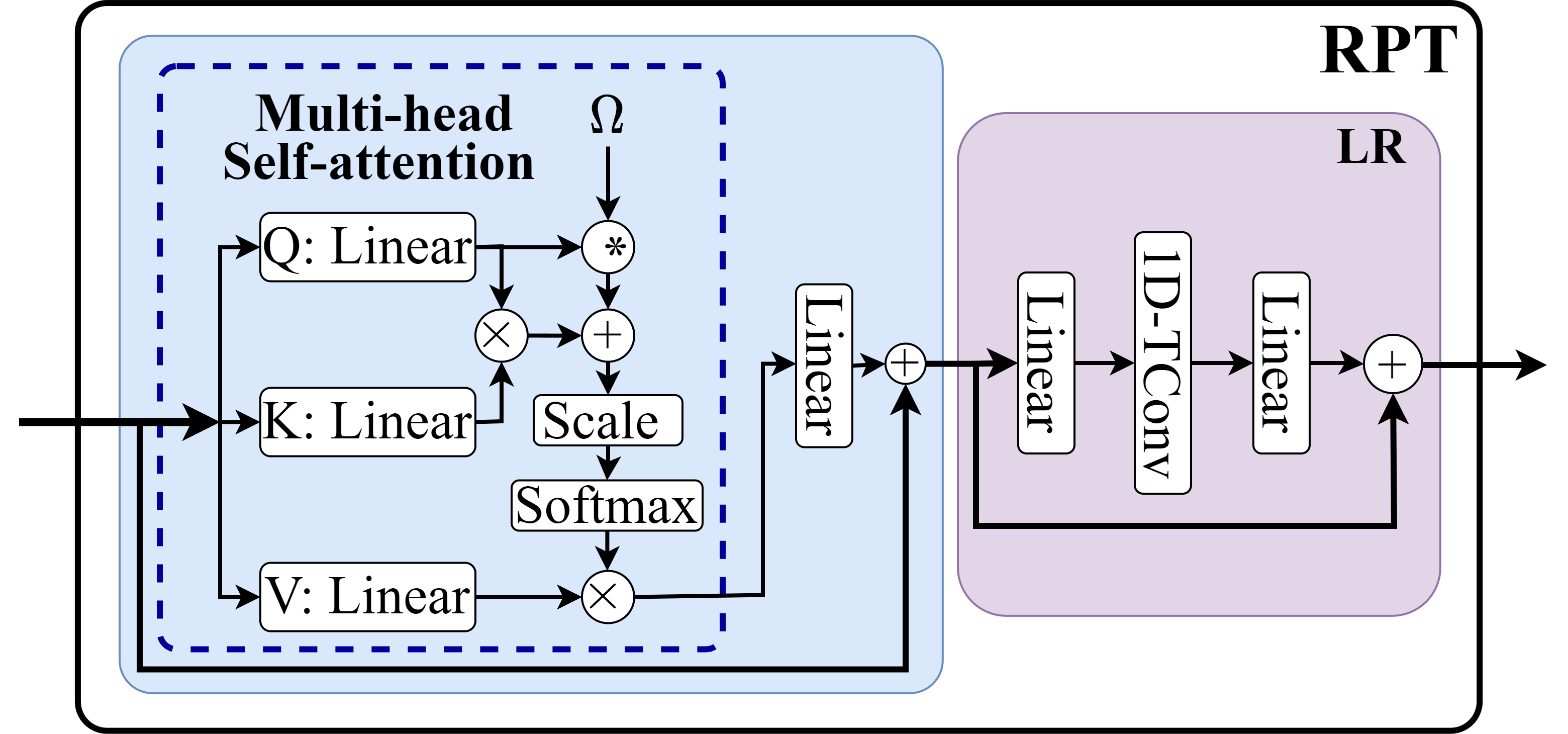}
  \caption{Architecture of {our proposed} relative positional transformer RPT block. An RPT block consists of a multi-head self-attention layer with the relative positional embedding followed by a local relational LR component. For brevity, the computation of the heads are not shown separately.}
  \label{fig:RPT}
\end{figure}

To design the ML-Rel, Fine-Det, and Coarse-Det modules, we utilize our proposed relative positional transformer (RPT) block, as illustrated in Fig. \ref{fig:RPT}. The RPT block consists of a transformer layer with relative positional embedding, followed by a local relational (LR) component. The LR component (\cite{dai2022ms}) includes two linear layers and one 1D temporal convolutional layer to enhance the output of the transformer layer.

We briefly formulate the transformer layer in our RPT block. In the H-head self-attention layer of RPT, for each head $h \in \{1, 2, ..., H\}$, the input sequence $X\in \rm {I\!R}^{N\times D^{\diamond}}$ is first transferred into query ${Q_h}$, key ${K_h}$, and value ${V_h}$ through linear operations
\begin{equation}
{Q}_{h} = XW_h^q,~~ {K}_{h} = XW^{k}_{h},~~{V}_{h} = XW^{v}_{h},
\end{equation}
where ${Q_h}$, ${K_h}$, ${V_h}\in \rm {I\!R}^{N\times D_h}$, $W_h^q,~W_h^k,~W_h^v\in \rm {I\!R}^{D^{\diamond}\times D_h}$ refer the  weights of linear operations, and $D_h = \frac{D^{\diamond}}{H}$. Then, the self-attention with relative positional embedding is computed for each head as 
\begin{equation}
{A}_h = softmax(\frac{{Q}_h {K}^T_h+{P}^{\triangleright}_h}{\sqrt{D_h}}){V}_h,
\end{equation}
\begin{equation}
{P}^{\triangleright}_h(n,m) = \sum_{d=1}^{D_{h}} {Q}_h(n, d)\Omega_{d}(n-m),
\end{equation}
where ${{P}^{\triangleright}_h}\in \rm {I\!R}^{N\times N}$, $n,m \in\{1, 2, ..., N\}$, and $\Omega_{d}$ operates as $D_{h}$ different embeddings for time intervals based on the queries (\cite{shaw2018self}). To compute ${{P}^{\triangleright}_h}$, we use the memory-efficient method proposed by \citet{huang2018music}.

Finally, the self-attention of all heads are concatenated and fed into a linear layer to output sequence ${O}$
\begin{equation}
{A} = concat(A_1, A_2,...,A_m),
\end{equation}
\begin{equation}
{O} = {A}W^{o}+X,
\end{equation}
where ${A}\in \rm {I\!R}^{N\times D^\diamond}$ and $W^{o}\in \rm {I\!R}^{D^\diamond\times D^\diamond}$.

\vspace{3mm}
\noindent {\bf Multi-Label Relationship Module (ML-Rel) -} The ML-Rel module encodes the co-occurrence action class relationships from the video’s ground-truth labels $G\in \rm {I\!R}^{T\times C}$. It contains a 1D temporal convolutional layer followed by $B$ RPT blocks. The convolution layer ${\Delta}$ has a kernel size of three and a stride of one to map $G$ into a different dimension $C^{*}$, and the RPT blocks are applied to exploit the temporal co-occurrence action dependencies amongst them    
\begin{equation}
{\hat{G}} = RPT^{MLRM}_{1:B}(\Delta({G})),
\end{equation}
where $\hat{G}\in \rm {I\!R}^{T\times C^*}$.

\vspace{2mm}
\noindent {\bf Multi-Label Classification Module (ML-CLAS) -} This module has a 1D convolution filter $CLAS^{\boxtimes}$ with kernel size one and stride one that is applied on the output of the ML-Rel module to decode the learned action class relationships into $C$ action class probabilities for each temporal moment
\begin{equation}
\label{eq:class probability 1}
Y^{\circledast} = Sig(CLAS^{\boxtimes}(\hat{G})),
\end{equation}
where $Y^{\circledast}\in \rm {I\!R}^{T\times C}$ and $Sig$ refers to sigmoid activation function.

\vspace{2mm}
\noindent {\bf Video Encoder (Vid-Enc) -} To process an input video, we need to convert it into a sequence of tokens. To perform this, similar to the previous action detection approaches (\cite{tirupattur2021modeling, dai2022ms, zhang2022actionformer}), we first divide the L-frame input video $V\in \rm {I\!R}^{L\times Ch\times W \times H}$ into T non-overlapped segments $S=\{S_t\}^{T}_{t=1}$, where ${S_t}\in \rm {I\!R}^{Z\times Ch\times W \times H}$, $Z=L/T$, and $Ch$, $W$, and $H$ define number of channels, width, and height of each video frame respectively. Then, Vid-Enc that is a pre-trained convolutional network is employed on each segment to generate its corresponding token 
\begin{equation}
    I_t = \text{Vid-Enc}(S_t),
\end{equation}
where $I_t\in \rm {I\!R}^{D}$.

\vspace{2mm}
\noindent {\bf Fine Detection Module (Fine-Det) -} The Fine-Det module aims to obtain a fine-grained temporal action dependency representation of the video from the input video sequence for the Coarse-Det and Vid-CLAS modules. The Fine-Det module includes a 1D temporal convolutional layer followed by $B$ RPT blocks. The convolution layer ${\Lambda^{\bullet}}$ has a kernel size of three and a stride of one to map all the input tokens $I\in \rm {I\!R}^{T\times D}$ into a lower dimension $D^{*}$, and then the RPT blocks are applied to learn the fine-grained dependencies $I^{\odot}$.  
\begin{equation}
{I}^{\odot} = RPT^{\text{Fine-Det}}_{1:B}({\Lambda^{\bullet}}(I)),
\end{equation}
where $I^{\odot}\in \rm {I\!R}^{T\times D^*}$ and $D^* < D$.

\begin{figure}[t]
  \centering
  \includegraphics[width=1.0\linewidth]{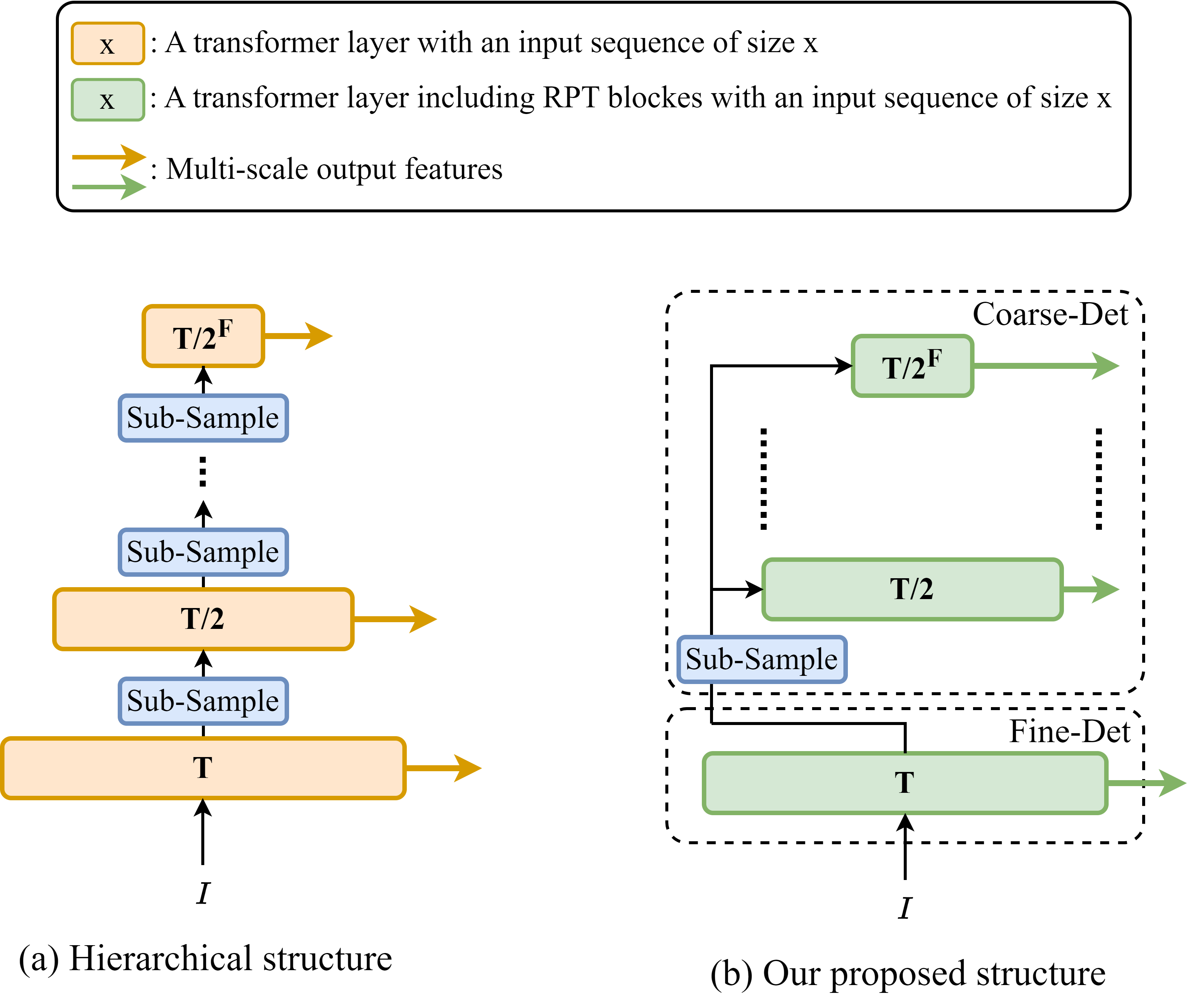}

  \caption{The proposed hierarchical structure in (\cite{dai2022ms, zhang2022actionformer}) vs. our proposed non-hierarchical design in the Fine-Det and Coarse-Det modules to extract multi-scale features for action detection.}
  \label{fig:hierachical vs ours}
\end{figure}

\begin{figure*}[t]
  \centering
  \includegraphics[width=0.8\linewidth]{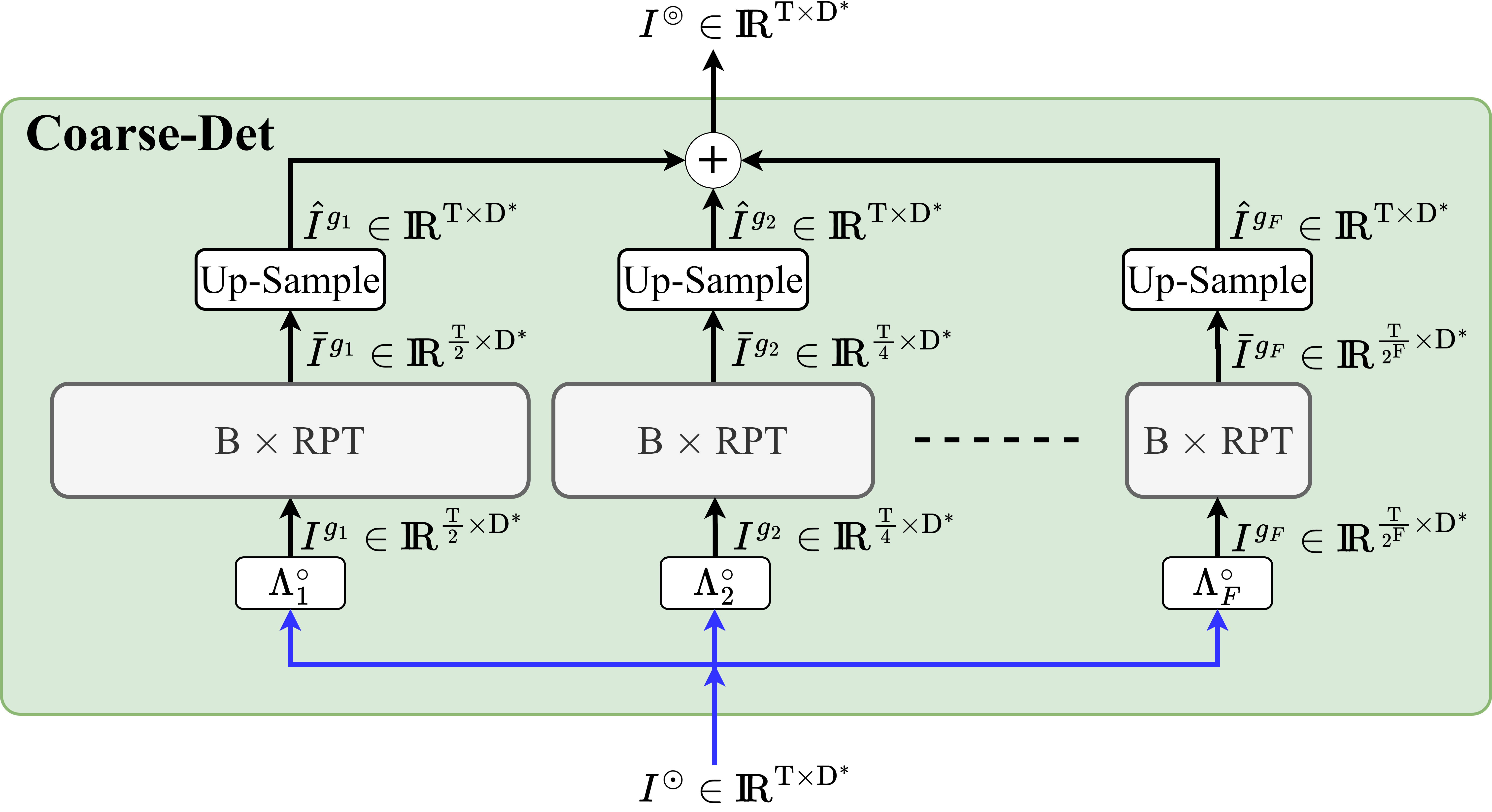}

  \caption{Architecture of {our proposed} Coarse-Det module that includes $F$ granularity branches to exploit different scales of temporal dependencies from the fine-grained features.}
  \label{fig: cdm}
\end{figure*}

\vspace{2mm}
\noindent {\bf Coarse Detection Module (Coarse-Det) -} In the Coarse-Det module, we aim to learn a coarse temporal action dependency representation of the video. To achieve this, one solution is to extract multi-scale temporal features through a hierarchical transformer-based structure, such as the proposed method in (\cite{dai2022ms, zhang2022actionformer}) (see Fig. \ref{fig:hierachical vs ours}. a). However, using multiple sub-sampling processes in the hierarchical structure results in losing positional information, specifically in the top layers of the network. Our Coarse-Det module has been designed to overcome this issue by extracting different scales of features from the same full-scale fine-grained information and through only one sub-sampling process, {(see Fig. \ref{fig:hierachical vs ours}. b)}. {In Section \ref{sec:ablation}, we show that our novel non-hierarchical design to extract multi-scale features outperforms significantly a hierarchical structure.} 

The Coarse-Det module comprises {$F$ granularity branches}, each designed to learn a distinct scale of temporal video features (Fig. \ref{fig: cdm}). Within the i-th branch, a 1D temporal convolutional layer $\Lambda^{\circ}_{i}$ with a kernel size of 3 and a stride of $2^i$, initially downsamples the fine-grained features received from the preceding Fine-Det module. The downsampled features then are given into $B$ RPT transformer blocks to effectively capture and model the temporal dependencies among them.
\begin{equation}
{I}^{g_i} = \Lambda^{\circ}_{i}({I}^{\odot}),
\end{equation}
\begin{equation}
\bar{I}^{g_i} = RPT^{\text{Coarse-Det}_{i}}_{1:B}({I}^{g_i}),
\end{equation}
where ${I}^{g_i}, {\bar{I}}^{g_i} \in \rm {I\!R}^{T^i\times D^*}$, $i\in\{1, 2, ..., F\}$, and $T^i = \frac{T}{2^i}$.

It is important to note that in our non-hierarchical structure, all feature scales are extracted from the same fine-grained information using a single striding (sub-sampling) process. Furthermore, since relative positional information is already embedded within the fine-grained features, the sub-sampled features retain their temporal positional cues even after the striding.

In the video classification module Vid-CLAS, we need to estimate action class probabilities for each input token generated by Vid-Enc. Therefore, the Coarse-Det module requires obtaining a coarse dependency representation at the original temporal length. To do this, we up-sample and combine different scales of coarse features to provide the final coarse representation
\begin{equation}
I^{\circledcirc} = \sum_{i=1}^{F} UpSample(\bar{I}^{g_i}),
\end{equation}
where $I^{\circledcirc}\in \rm {I\!R}^{T\times D^*}$ and linear interpolation is employed for up-sampling.
  
\vspace{2mm}
\noindent {\bf Video Classification Module (Vid-CLAS) -} 
This module has been designed to achieve two goals: (i) estimating action class probabilities from both fine and coarse contexts for the action detection task, and (ii) encouraging the fine and coarse detection modules to learn the same co-occurrence action class  dependencies as those learned from the ground-truth labels in the multi-label relationships module of the Assistant branch.

In previous approaches, \citet{zhang2022actionformer} predict action class probabilities for each temporal scale of their hierarchical structures. Alternatively, \citet{dai2022ms} first fused all scales, {\ie}, from fine to coarse, and then predicted class probabilities for the fused features. However, we observed that in our proposed network, the best results are achieved by obtaining the action class probabilities from the fine and coarse action representations. To accomplish this, the Vid-CLAS module employs a 1D convolutional filter $CLAS^{\otimes}$ which is applied on both the fine and coarse features separately to predict $C$ action class probabilities for each temporal moment
\begin{equation}
\label{eq:class probability 2}
Y^{\phi} = Sig(CLAS^{\otimes}(I^{\phi})),
\end{equation}
where $Y^{\phi}\in \rm {I\!R}^{T\times C}$ and $\phi\in\{\odot, \circledcirc\}$. At the inference, the final estimation is computed by combining the action predictions obtained from both fine and coarse features as 
\begin{equation}
{Y} = {\sum_{\phi}\alpha_{\phi}Y^{\phi}},
\end{equation}
where ${Y}\in \rm {I\!R}^{T\times C}$ and $\alpha_{\odot}+~\alpha_{\circledcirc} = 1$. The $CLAS^{\otimes}$ filter employs the same architecture as $CLAS^{\boxtimes}$ in the ML-Rel module of the Assistant branch, and copies its parameters,
\begin{equation}
    CLAS^{\otimes}_{\Theta} \longleftarrow CLAS^{\boxtimes}_{\Theta}.
\end{equation}
The $CLAS^{\otimes}$'s parameters are frozen during the training. This encourages the Core branch to capture the same cross-class action dependencies in the fine and coarse detection modules as those learned from the ground-truth labels in the Assistant branch.



\subsubsection{Network Optimization} 
Action detection approches (\cite{tirupattur2021modeling, dai2022ms, zhang2022actionformer, vahdani2022deep})  usually employ the binary cross entropy (BCE) loss function for optimization. However, in the multi-label setting, the number of positive labels may become more than the number of negative ones. This unbalanced number of positive and negative labels can result in poor performance in the action detection task if we employ BCE for training, since it does not have any control on the contribution of positive and negative samples. To overcome this, {we propose to adapt Asymmetric loss $\mathcal{L}_{asl}$ (\cite{ridnik2021asymmetric}) for multi-label action detection task}. Therefore, the Assistant and Core branches are optimized through the $\mathcal{L}_{Assis}$ and $\mathcal{L}_{Core}$ losses, respectively, as
\begin{equation}
\mathcal{L}_{Assis} = \frac{1}{T}\sum_{t=1}^{T}\sum_{c=1}^{C}\mathcal{L}_{asl}({g}_{t,c}, {y}^{\circledast}_{t,c}),
\end{equation}
\begin{equation}
\mathcal{L}_{Core} = \frac{1}{T}{\sum_{\phi}}\sum_{t=1}^{T}\sum_{c=1}^{C}\alpha_{\phi}\mathcal{L}_{asl}({g}_{t,c}, {y}^{\phi}_{t,c}),
\end{equation}
where ${g}_{t,c}$ indicates the ground truth label of action class $c$ in temporal step $t$, and ${y}^{\circledast}_{t,c}$ and ${y}^{\phi}_{t,c}$ are its corresponding class probability estimated by Eqs. \ref{eq:class probability 1} and   \ref{eq:class probability 2} respectively, and 
\begin{equation}
\mathcal{L}_{asl}({g}, {y}) = -{g}\mathcal{L}_{+}-(1-{g})\mathcal{L}_{-},
\end{equation}
\begin{equation}
\label{eq:focus+}
\mathcal{L}_{+} = (1-{y})^{\gamma_+}log({y}),~ \mathcal{L}_{-} = {{y}}^{\gamma_-}log(1-{y}),
\end{equation}
where $\gamma_+$ and $\gamma_-$ are focusing parameters for positive and negative labels respectively. 

\section{Experimental Results}
\label{sec:experiments}
\noindent {\bf Datasets --} There are several benchmark datasets for action detection, but only a few of them provide dense multi-label annotations. For instance, videos in ActivityNet (\cite{activitynet}) have only one action type per timestamp. We present the results of our proposed approach on two challenging dense multi-label benchmark datasets, {\char} (\cite{charades}) and {\thum} (\cite{multithomus}). 

{\char} is a large dataset including $9,848$ videos of daily activities of 267 persons. It contains $66,500$ temporal interval annotations for 157 action classes while there is a high overlap amongst the action instances of different action categories. To evaluate our method on {\char}, we follow previous methods (\cite{kahatapitiya2021coarse,tirupattur2021modeling,dai2022ms}) and use the same training and testing set as in (\cite{charades}).

{\thum} contains the same set of 413 videos as in THUMOS’14 dataset (\cite{jiang2014thumos}). However, \noindent {\thum} is more challenging than THUMOS’14 since (i) the annotations have been extended from 20 action classes to 65, and (ii) in contrast to sparse-label frame-level annotations in THUMOS’14, {\thum} has dense multi-label action annotations. To obtain the results on this dataset, we use the same standard training and testing splits applied by previous methods \cite{tirupattur2021modeling, dai2022ms}.

\vspace{2mm}
\noindent {\bf Implementation Details --} Similar to the proposed method in (\cite{dai2022ms}), our network uses a fixed number of $T=256$ input tokens during both training and inference. For training, we randomly sample a clip containing $T$ consecutive tokens from a video sequence. At inference, we follow previous methods (\cite{tirupattur2021modeling,kahatapitiya2021coarse}) and make the predictions for the full video sequence. Each input token is provided by applying the video encoder Vid-Enc on an 8-frame segment to extract a feature vector with dimension $D = 1024$. Vid-Enc is implemented using a pre-trained I3D\footnote{{Video encoder Vid-Enc is pre-trained on Kinetic-400 (\cite{kay2017kinetics}) and training set of {\char} for {\thum} and {\char} respectively}.} (\cite{carreira2017quo}) while its fully connected layers are replaced with a global average pooling layer {and its parameters are frozen}. In the convolutional layer of the ML-Rel and Fine-Det modules, the input features are mapped into $C^{*}=D^{*}=512$ dimensional feature vectors. 
The ML-Rel and the Fine-Det modules, as well as {each granularity branch} of the Coarse-Det module, have $B=3$ RPT blocks with $H=8$ multi-head attention heads. 
Table \ref{tab:details of mf-vit} shows details of our proposed network architecture. The contributing factors for fine-grained ($\alpha_{\odot}$) and coarse-grained ($\alpha_{\circledcirc}$) features in the V-CLAS module are set {empirically} to $\{\alpha_{\odot}=0.1, \alpha_{\circledcirc}=0.9\}$ and $\{\alpha_{\odot}=0.7, \alpha_{\circledcirc}=0.3\}$ for {\char} and {\thum} respectively. In Asymmetric loss,
we use factors of $\gamma_+=1$ and $\gamma_-=3$ for the impact of positive and negative samples respectively, which are determined through trial and error. 

Our experiments were performed under Pytorch on an NVIDIA GeForce RTX 3090 GPU, and we trained our model using the Adam optimiser with an initial learning rate of 0.0001 and batch size 3 for 25 and 300 epochs for {\char} and {\thum} datasets respectively. The learning rate was decreased by a factor of 10 every 7 and 130 epochs for {\char} and {\thum} respectively. {Note, using different training settings for {\char} and {\thum} is due to their different size.}

\begin{table}[h]

  \centering
  \setlength{\tabcolsep}{6pt} 
    \renewcommand{\arraystretch}{1.2} 
  \begin{tabular}{@{}l @{}}\specialrule{.2em}{.1em}{.1em}
    \multicolumn{1}{c}{Details of Modules}\\\specialrule{.2em}{.1em}{.1em}
    {\bf ML-Rel} \\ 
    ~~~~$Conv(512, 3, 1)$, {\bcolor{$Nrm$}, $3\times RPT$}\\\midrule
    {\bf {{ML-CLAS}}}\\
    ~~~~$CLAS^{\boxtimes}:Conv(C, 1, 1)$\\\midrule
    {\bf Fine-Det} \\ 
    ~~~~$Conv(512, 3, 1)$, {\bcolor{$Nrm$}, $3\times RPT$}\\\midrule
    {\bf {Coarse-Det}}\\
    ~~~~{{Granularity Branch 1}: $Conv(512, 3, 2)$, {\bcolor{$Nrm$}}, $3\times RPT$}\\
    ~~~~{{Granularity Branch 2}: $Conv(512, 3, 4)$, {\bcolor{$Nrm$}}, $3\times RPT$}\\
    ~~~~{{Granularity Branch 3}: $Conv(512, 3, 8)$, {\bcolor{$Nrm$}}, $3\times RPT$}\\\midrule
    {\bf {{Vid-CLAS}}}\\
    ~~~~$CLAS^{\otimes}:Conv(C, 1, 1)$\\\midrule
    {\bf {RPT}} \\
    {Self-attention -- }\\
    ~~~~Query:{\bcolor{$Nrm$}}, $Linear(512)$,\\
    ~~~~Key:{\bcolor{$Nrm$}}, $Linear(512)$,\\
    ~~~~Value:{\bcolor{$Nrm$}}, $Linear(512)$,\\
    ~~~~$Linear(512)$\\
    {RL --} \\
    ~~~~{\bcolor{$Nrm$}}, $Linear(512)$, $Conv(512, 3, 1)$, \\
    ~~~~\bcolor{$Drp$}, $Linear(512)$, \bcolor{$Drp$} \\\specialrule{.2em}{.1em}{.1em}
  \end{tabular}
  \caption{Details of our proposed network's architecture. Conv(ch, k, s): 1D temporal convolution filter with ch output channel, a kernel size of $k$, and a stride of size $s$. Linear(d): a linear layer with the output size of $d$. Nrm: normalisation layer. Drp: dropout. C is maximum number of action classes in a dataset. \vspace{-10mm}} 
  \label{tab:details of mf-vit}
\end{table}

\vspace{2mm}
\subsection{Ablation Studies}
\label{sec:ablation}
In this section, we examine our design decisions for the proposed network and learning paradigm.

\vspace{2mm}
\noindent{\bf Number of Granularity Branches in Coarse-Det Module -- } To design the Coarse-Det module, we experimented with different numbers of granularity branches, from one to four for which their results are available in Table \ref{tab:number of branches}. The results show that the best performance is achieved when the number of branches is set to 3. Therefore, we used a Coarse-Det module with 3 granularity branches to perform all experiments.

\begin{table}[t]
  \centering
  \setlength{\tabcolsep}{4pt} 
    \renewcommand{\arraystretch}{1.0} 
  \begin{tabular}{@{}c c c c @{}}\specialrule{.2em}{.1em}{.1em}
    \multirow{2}{*}{{~~~\# Branch}~~~} & ~~~~~~~~~& \multicolumn{2}{c}{{mAP(\%)}}\\\cmidrule{3-4}
    && {\char} & {\thum}\\ \specialrule{.2em}{.1em}{.1em}
   {1} && {24.5}& 43.0\\
   {2} && 25.8 & 43.9\\ 
   {3} && \bf 26.5& \bf 44.6\\ 
   {4} && 25.9& 44.2\\ \specialrule{.2em}{.1em}{.1em}
  \end{tabular}
  \caption{Ablation studies on the number of granularity branches in the Coarse-Det module of our method on the {\char} and {\thum} datasets using RGB videos in terms of per-frame mAP metric.} 
  \label{tab:number of branches}
\end{table}

\begin{table}[t]
\scalebox{1.0}
{
  \setlength{\tabcolsep}{3pt} 
    \renewcommand{\arraystretch}{1.2} 
  \begin{tabular}{@{}c c c c c @{}}\specialrule{.2em}{.1em}{.1em}
    \multirow{2}{*}{Fine-Det}&\multirow{2}{*}{Coarse-Det}&& \multicolumn{2}{c}{{mAP(\%)}} \\ \cmidrule{4-5}
    & & & {\char}&{\thum}\\ \specialrule{.2em}{.1em}{.1em}
   {$\times$}& {$\times$}& & 16.8 & 36.7 \\
   {\checkmark}& {$\times$}  & & 23.8  & 40.5\\
   {$\times$}& {\checkmark} & & 26.2 & 40.1\\  \cmidrule{1-5}
   {\checkmark}& {\checkmark} &  & \bf 26.5 &\bf 44.6\\ \specialrule{.2em}{.1em}{.1em}
  \end{tabular}}
  \caption{Ablation studies on the Fine-Det and Coarse-Det modules of our proposed approach on the {\char} and {\thum} dataset using RGB videos in terms of per-frame mAP metric.}
  \label{tab: fine and coarse}
\end{table}

\vspace{2mm}
\noindent{\bf Effect of Fine-Det Module \& Coarse-Det Module --}
Here, we evaluate the impact of the Fine-Det and the Coarse-Det modules in final results of our method by removing each or both of the modules. To obtain the results of the network when both modules are dropped, we use directly the sequence of input tokens generated by Vid-Enc (I3D) for action detection. Table \ref{tab: fine and coarse} shows that using only input tokens generated by Vid-Enc (I3D) is not enough for effective action detection and employing fine and coarse-grained temporal features obtained by Fine-Det and Coarse-Det improves the performance by $9.7\%$ and $7.9\%$ per-frame mAP on {\char} and {\thum}, respectively. It also shows that both Fine-Det and Coarse-Det modules have an important contribution to the final results as by removing them, our results deteriorate by $2.4\%$ and $3.4\%$ per-frame mAP on average on both datasets, respectively.

\vspace{2mm}
\noindent{\bf What Should be the Input of Coarse-Det Module?} In our proposed approach, the Coarse-Det module receives the output of the Fine-Det module as input to extract its features. This is because we found that this network design outperforms a design where the coarse or multi-glandular features are extracted directly from the input tokens. We present the results of both designs in Tabel \ref{tab: str}.

\begin{table}[t]
  \centering
  \setlength{\tabcolsep}{4pt} 
    \renewcommand{\arraystretch}{1.2} 
  \begin{tabular}{@{}l c c c @{}}\specialrule{.2em}{.1em}{.1em}
    \multirow{2}{*}{{Design}} &~~~~~~~~~~~~~~~~~~~~& \multicolumn{2}{c}{{mAP(\%)}}\\\cmidrule{3-4}
    & & {\char} & {\thum}\\ \specialrule{.2em}{.1em}{.1em}
   {Net-$v_1$} && {26.1}& 44.2\\ \midrule
   Ours & & \bf 26.5& \bf 44.6\\ \specialrule{.2em}{.1em}{.1em}
  \end{tabular}
  \caption{Ablation studies on the input of the Coarse-Det module on the {\thum} and  {\char} datasets using RGB videos in terms of per-frame mAP metric. In Net-$v_1$, the Coarse-Det module directly uses the input tokens extracted by Vid-Enc as input, while the rest of the network remains the same as in our proposed approach.}
  \label{tab: str}
\end{table}

\vspace{2mm}
\noindent{\bf Effect of Assistant Branch \& Our Learning Paradigm --} 
Here, we aim to evaluate the impact of the Assistant branch that is employed by our proposed learning paradigm. Table \ref{tab:mf-vit components} compares the performance of our proposed method with and without using the Assistant branch during the training. Note, to obtain the results of the network when the Assistant branch is omitted, the Vid-CLAS's parameters are not frozen and take part in the training. Table \ref{tab:mf-vit components} shows that adding the Assistance branch and using our new learning paradigm boosts the method’s performance by $0.5\%$ per-frame mAP on both datasets. Importantly, our method benefits from this improvement during inference without suffering from the computational overhead associated with explicitly modeling action relationships.

\begin{table}[t]
\begin{center}
\scalebox{0.85}
{
  \setlength{\tabcolsep}{4pt} 
    \renewcommand{\arraystretch}{1.5} 
  \begin{tabular}{@{}c c c c c @{}}\specialrule{.2em}{.1em}{.1em}
    \multicolumn{2}{c}{Branch}&~~~~~~~~~~~~~~& \multicolumn{2}{c}{{mAP(\%)}} \\ \cmidrule{1-2}\cmidrule{4-5}
    Core & Assistant & ~ & {\char}&{\thum}\\ \specialrule{.2em}{.1em}{.1em}
   {\checkmark}& {$\times$}& & 26.0 & 44.1\\ \cmidrule{1-5}
   {\checkmark}& {\checkmark}& & \bf 26.5 & \bf 44.6\\ \specialrule{.2em}{.1em}{.1em}
  \end{tabular}}
  \caption{Ablation studies on the network branch of our proposed approach employed during the training stage on the {\char} and {\thum} dataset using RGB videos in terms of per-frame mAP metric.}
  \label{tab:mf-vit components}
  \end{center}
\end{table}

\begin{table}[t]
  \centering
  \setlength{\tabcolsep}{6pt} 
    \renewcommand{\arraystretch}{1.0} 
  \begin{tabular}{@{}l c c @{}}\specialrule{.2em}{.1em}{.1em}
    \multirow{2}{*}{{Design}}& \multicolumn{2}{c}{{mAP(\%)}}\\\cmidrule{2-3}
    & {\char} & {\thum}\\ \specialrule{.2em}{.1em}{.1em}
    {Net-$v_2$ (Hierarchical)} & 25.1 & 44.0\\ \midrule
   Ours & \bf 26.5& \bf 44.6\\ \specialrule{.2em}{.1em}{.1em}
  \end{tabular}
  \caption{Ablation studies on structure design of the proposed method on the {\char} and {\thum} datasets using RGB videos in terms of per-frame mAP metric. In Net-$v_2$, the Coarse-Det module employees a hierarchical structure to learn temporal dependencies, while the rest of the network remains the same as in our proposed approach.} 
  \label{tab:different structures}
\end{table}

\vspace{2mm}
\noindent{\bf Effect of Our Proposed Non-hierarchical Structure --}
In this section, we compare the design of our proposed network with a variant of it, namely, Net-$v_2$. In Net-$v_1$, the Coarse-Det module utilizes a hierarchical structure to extract multi-scale features, while the rest of its architecture remains identical to our proposed network. Table \ref{tab:different structures} shows that when the Coarse-Det module applies a hierarchical structure to learn the coarse-grained features, {\ie} {Net-$v_1$}, the method's performance drops by $1.4\%$ and {$0.6\%$} perf-frame mAP on {\char} and {\thum}, respectively. {This proves the contribution of our novel non-hierarchical transformer-based design which preserves positional information when exploiting the multi-scale temporal features.}


\vspace{2mm}
\noindent {\bf {Effect of Relative Positional Encoding --}} Table \ref{tab:positional encoding} shows the performance of our network when different positional encodings are applied. It can be observed that employing the relative positional encoding (\cite{shaw2018self,huang2018music}) embedded in our RPT block improves the method's performance by $0.3\%$ per-frame mAP on both datasets, while adding absolute positional encoding (\cite{vaswani2017attention}) into the input tokens deteriorates the method's performance significantly.

\begin{table}[h]
\scalebox{1.0}
{
  \setlength{\tabcolsep}{6pt} 
    \renewcommand{\arraystretch}{1.0} 
  \begin{tabular}{@{}l c c @{}}\specialrule{.2em}{.1em}{.1em}
    \multirow{2}{*}{{Positional Encoding}}& \multicolumn{2}{c}{{mAP(\%)}}\\\cmidrule{2-3}
    & {\char} & {\thum}\\ \specialrule{.2em}{.1em}{.1em}
   {No encoding} & 26.2& 44.3\\
   {Absolute} & 25.3 & 43.5\\\midrule
   {Relative} & \bf 26.5& \bf 44.6\\ \specialrule{.2em}{.1em}{.1em}
  \end{tabular}}
  \caption{Ablation studies on positional encoding used in our proposed network on the {\char} and {\thum} dataset using RGB videos in terms of per-frame mAP metric.}
  \label{tab:positional encoding}
\end{table}

\vspace{2mm}
\noindent{\bf Effect of Loss Function --}
Here, we examine the effect of BCE and Asymmetric (\cite{ridnik2021asymmetric}) losses for training. As shown in Table \ref{tab:different losses}, applying the Asymmetric loss to optimize our network improves the performance by $0.5\%$ and $0.2\%$ per-frame mAP on {\char} and {\thum} respectively.

\begin{table}[t]
  \centering
  \setlength{\tabcolsep}{6pt} 
    \renewcommand{\arraystretch}{1.0} 
  \begin{tabular}{@{}l c c @{}}\specialrule{.2em}{.1em}{.1em}
    \multirow{2}{*}{{Loss}}& \multicolumn{2}{c}{{mAP(\%)}}\\\cmidrule{2-3}
    & {\char} & {\thum}\\ \specialrule{.2em}{.1em}{.1em}
    {BCE} & {26.0}& 44.4\\ \midrule
   Asymmetric & \bf 26.5& \bf 44.6\\ \specialrule{.2em}{.1em}{.1em}
  \end{tabular}
  \caption{Ablation studies on the loss function applied for training our network on the {\char} and {\thum} datasets using RGB videos in terms of per-frame mAP metric.} 
  \label{tab:different losses}
\end{table}

\vspace{2mm}
\noindent {\bf Discussion and Analysis --} {The ablation studies show that leveraging positional information in the transformer layers has an important contribution in the final results of the network where extracting the multi-scale temporal features through our novel non-hierarchical design in Coarse-Det outperforms a hierarchical structure by $1.0\%$ per-frame mAP on average on both datasets, and embedding the relative position encoding in the RPT block improves the performance by $0.3\%$ per-frame mAP on both datasets. Furthermore, the experiments showed that the Core branch learns the action dependencies more effectively when it is trained through our proposed learning paradigm and benefits from learned knowledge in the Assistant branch, \ie the method’s performance is boosted by $0.5\%$ per-frame mAP on both datasets. }



\begin{table*}[t]
\begin{center}
\scalebox{0.95}
{
  \centering
  \setlength{\tabcolsep}{4pt} 
    \renewcommand{\arraystretch}{0.9} 
  \begin{tabular}{@{}l c c c c@{}}\specialrule{.2em}{.1em}{.1em}
    \multicolumn{1}{l}{\multirow{2}{*}{Method}}& \multirow{2}{*}{GFLOPs} & \multicolumn{1}{c}{\multirow{2}{*}{Backbone}} & \multicolumn{2}{c}{mAP(\%)}\\ \cmidrule{4-5}
          & &  &\multicolumn{1}{c}{Charades} & \multicolumn{1}{c}{MultiTHUMOS}\\ \specialrule{.2em}{.1em}{.1em}
          R-C3D \cite{xu2017r}, ICCV&-& C3D & 12.7 & - \\ 
          TGM (\cite{piergiovanni2019temporal}), ICML&1.2&I3D &20.6&37.2\\ 
          {PDAN} {\checkmark}$*$ (\cite{dai2021pdan}), WACV&3.2&I3D &23.7&40.2\\ 
          CoarseFine (\cite{kahatapitiya2021coarse}), CVPR&-&X3D & 25.1&-\\ 
          MLAD {\checkmark}(\cite{tirupattur2021modeling}), CVPR&44.8 &I3D&18.4&42.2\\ 
          CTRN {\checkmark}(\cite{dai2021ctrn}), BMVC&-&I3D&25.3& {\underline{44.0}}\\
          PointTAD (\cite{tanpointtad}), NeurIPS&-&I3D& 21.0 & 39.8 \\
          MS-TCT {\checkmark}(\cite{dai2022ms}), CVPR&6.6&I3D &{\underline{25.4}}&{43.1}\\ \midrule
          {Ours} {\checkmark}&8.5& I3D&{\bf 26.5}&{\bf 44.6}\\\specialrule{.2em}{.1em}{.1em}
  \end{tabular}}
  \caption{Action detection results on {\char} and {\thum} datasets using RGB videos in terms of per-frame mAP. The {{\checkmark}} symbol highlights the transformer-based approaches, and $*$ indicates the results are taken from (\cite{dai2022ms}). The best and the second-best results are in {\bf Bold} and \underline{underlined}, respectively.} 
  \label{tab:sota v1}
  \end{center}
\end{table*}

\begin{table*}[t]
\begin{center}

  \centering
  \setlength{\tabcolsep}{3.0pt} 
    \renewcommand{\arraystretch}{1.0} 
      \begin{tabular}{@{}cl c c c c c @{}} \specialrule{.2em}{.1em}{.1em}
          &\multicolumn{1}{l}{\multirow{2}{*}{Method}}& &{~~~$P_{AC}$(\%)~~~}&{~~~$R_{AC}$~~~}&{~~~${F1}_{AC}$~~~}&{${mAP}_{AC}$}\\ 
          && &(\%)&{(\%)}&(\%)&{(\%)}\\ \specialrule{.2em}{.1em}{.1em}
          \multirow{5}{*}{~~\rotatebox{90}{$\tau = 0$~~~}~~}& I3D $*$ (\cite{carreira2017quo}), CVPR && 14.3 & 1.3 & 2.1 & 15.2 \\
          &CF $*$ (\cite{tirupattur2021modeling}), CVPR && 10.3 &1.0 &1.6 &15.8\\
          &MLAD {{\checkmark}} (\cite{tirupattur2021modeling}), CVPR &&19.3 &7.2 &8.9 &28.9\\
          &MS-TCT {{\checkmark}} (\cite{dai2022ms}), CVPR && \underline{26.3} & \underline{15.5} &\underline{19.5} &\underline{30.7}\\ \cmidrule{2-7}
          &{Ours} {{\checkmark}} && \bf 28.3 &\bf 26.1 &\bf 27.2 &\bf 32.0 \\ \specialrule{.2em}{.1em}{.1em}
          \multirow{5}{*}{~~\rotatebox{90}{$\tau = 20$~~~}~~}& I3D $*$ (\cite{carreira2017quo}), CVPR && 12.7 & 1.9 & 2.9 & 21.4  \\
          &CF $*$ (\cite{tirupattur2021modeling}), CVPR && 9.0&1.5&2.2&22.2\\
          &MLAD {{\checkmark}} (\cite{tirupattur2021modeling}), CVPR &&18.9& 8.9& 10.5& 35.7\\
          &MS-TCT {{\checkmark}} (\cite{dai2022ms}), CVPR && \underline{27.6}& \underline{18.4}& \underline{22.1}& \underline{37.6}\\ \cmidrule{2-7}
          &{Ours} {{\checkmark}} && \bf 30.0 &\bf 29.2 &\bf 29.6 &\bf 37.8 \\ \specialrule{.2em}{.1em}{.1em}
          \multirow{5}{*}{~~\rotatebox{90}{$\tau = 40$~~~}~~}& I3D $*$ (\cite{carreira2017quo}), CVPR && 14.9 & 2.0&3.1&20.3  \\
          &CF $*$ (\cite{tirupattur2021modeling}), CVPR && 10.7&1.6&2.4&21.0\\
          &MLAD {{\checkmark}} (\cite{tirupattur2021modeling}), CVPR &&19.6&9.0&10.8&34.8\\
          &MS-TCT {{\checkmark}} (\cite{dai2022ms}), CVPR && \underline{27.9} &\underline{18.3} &\underline{22.1} &\underline{36.4}\\ \cmidrule{2-7}
          &{Ours} {{\checkmark}} && \bf 30.0 &\bf 29.1 &\bf 29.4 &\bf 36.7  \\ \specialrule{.2em}{.1em}{.1em}
    \end{tabular}
  
  \caption{Action detection results on {\char} dataset based on the action-conditional metrics, {$P_{AC}$}, {$R_{AC}$}, {${F1}_{AC}$}, and {${mAP}_{AC}$}. {$\tau$} refers the temporal window size. Following previous approaches (\cite{dai2022ms,tirupattur2021modeling}), RGB and optical flow are used as input. The {{\checkmark}} symbol highlights the transformer-based approaches, and $*$ indicates the results are taken from (\cite{tirupattur2021modeling}). The best and the second-best results are in {\bf Bold} and \underline{underline}, respectively.}
  \label{tab:sota v2}
  \end{center}
\end{table*}

\subsection{State-of-the-Art Comparison} 
In this section, we compare the performance of the proposed method with state-of-the-art approaches. Table \ref{tab:sota v1} presents comparative results on the benchmark datasets {\char} and {\thum}, using the standard per-frame mAP metric. For a fair comparison, all transformer-based approaches employ a pre-trained, frozen I3D network to generate input tokens from input video images. Table \ref{tab:sota v1} shows that our proposed approach outperforms the current state-of-the-art result by $1.1\%$ and $0.6\%$ mAP on {\char} and {\thum}, respectively, and achieves a new state-of-the-art per-frame mAP results at $26.5\%$ and $44.6\%$ on {\char} and {\thum}, respectively. 


We also evaluate the performance of our proposed method by action-conditional metrics including Action-Conditional Precision {$P_{AC}$}, Action-Conditional Recall {$R_{AC}$}, Action-Conditional F1-Score {${F1}_{AC}$}, and Action-Conditional Mean Average Precision {${mAP}_{AC}$}, as introduced in (\cite{tirupattur2021modeling}). The aim of these metrics is to measure the ability of the network to learn both co-occurrence and temporal dependencies of different action classes. The metrics are measured throughout a temporal window with a size of $\tau$. As shown by the results {on {\char}} in Table \ref{tab:sota v2}, the proposed method {\method} achieves state-of-the-art results on all action-conditional metrics, specifically, it improves the state-of-the-art results significantly on $R_{AC}$ and ${F1}_{AC}$ by $10.6\%$ and $7.7\%$, $10.8\%$ and $7.5\%$, and $10.8\%$ and $7.3\%$ where $\tau$ is 0, 20, and 40 respectively.

\begin{figure}[h]
  \includegraphics[width=1.0\linewidth]{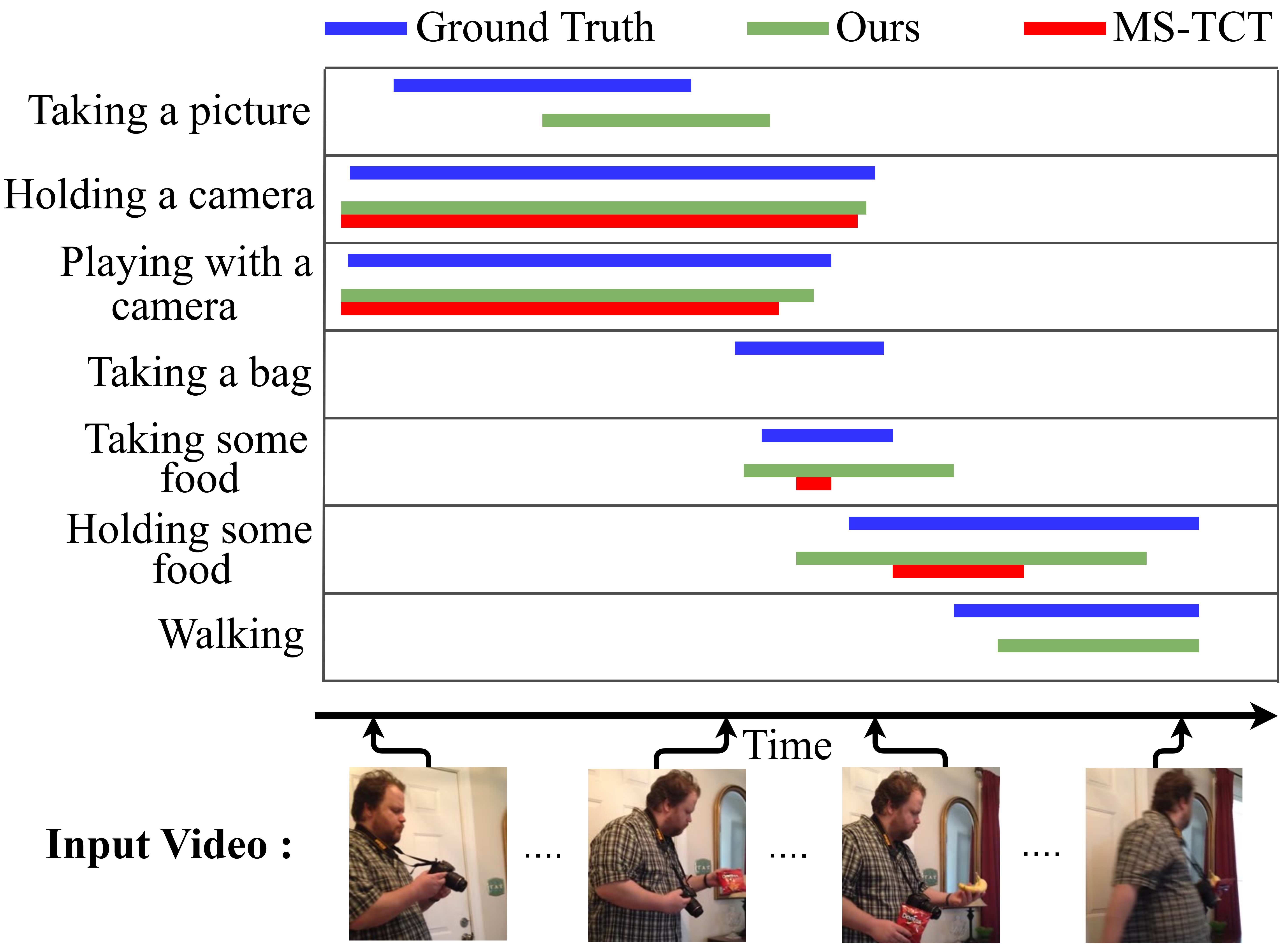}
  \caption{Visualization of action predictions by our proposed method and MS-TCT (\cite{dai2022ms}) on a test video sample of {\char} including 7 different action types.}
  \label{fig:qualitative}
\end{figure}

{Fig. \ref{fig:qualitative} displays qualitative results of {\method} on a test video sample of {\char} and compares them with the outputs of MS-TCT (\cite{dai2022ms}). Amongst the state-of-the-art methods, we applied MS-TCT (\cite{dai2022ms}) and MLAD (\cite{tirupattur2021modeling}) on the video sample, since their code is available, useable and compatible with our hardware. However, as the MLAD could not predict any of the actions, we reported only the results of MS-TCT. The results in Fig. \ref{fig:qualitative} show that our proposed method’s action predictions have a better overlap with the ground-truth labels, and our method detected more action instances in the video than MS-TCT, {\ie} {\method} predicted all action types except “{\it{Taking a bag}}” while MS-TCT could not detect “{\it{Taking a picture}}”, “{\it Taking a bag}”, and “{\it{Walking}}”.}

\section{Limitations}
\label{sec:limitations}
In our proposed network, the Vid-Enc encoder extracts the input tokens by a pre-trained model {that is not optimized with the rest of the network during the training process}. Thus, if the pre-trained model cannot encode the input images with strong features, the proposed method cannot achieve high performance. This limitation is observed in all the existing methods which use pre-trained models, such as \cite{tirupattur2021modeling, dai2022ms, zhang2022actionformer}.

\section{Conclusion}
\label{sec:conclusion}
In this work, we introduced a novel transformer-based network that is trained through a novel learning paradigms to effectively and efficiently learn complex temporal action dependencies for dense multi-label action detection. Preserving temporal positional information is essential for accurate action detection, and our network is designed to leverage this important information by employing a non-hierarchical structure when exploiting temporal dependences, and embedding relative positional information in its transformer layers. Our proposed learning paradigm enabels the network to benefit from explicit modelling temporal cross-action relations obtained from the ground-truth video’s labels during the training without adding their associated computational costs at the inference.  We evaluated our approach on two densely-labelled challenging benchmark action detection datasets, on which we achieved new state-of-the-art results, and our ablation studies demonstrated the contributions of different components of our network and the effectiveness of the our learning paradigm. For future work, we will investigate adapting our network to learn spatial and temporal dependencies from raw pixels and also use audio information to improve the performance of action detection.

\section*{Acknowledgement}
This research is supported by UKRI EPSRC Platform Grant EP/P022529/1, and EPSRC BBC Prosperity Partnership AI4ME: Future Personalised Object-Based Media Experiences Delivered at Scale Anywhere EP/V038087/1.

\section*{Data Availability Statement}
The {\char} and {\thum} datasets used in this paper are available at \href{https://prior.allenai.org/projects/charades}{https://prior.allenai.org/projects/charades} and \href{https://ai.stanford.edu/~syyeung/everymoment.html}{https://ai.stanford.edu/syyeung/everymoment.html}, respectively.







\bibliography{sn-bibliography}

\end{document}